# Approximate Muscle Guided Beam Search for Three-Index Assignment Problem

He Jiang, Shuwei Zhang, Zhilei Ren, Xiaochen Lai, and Yong Piao

Software School, Dalian University of Technology, Dalian, 116621, China
jianghe@dlut.edu.cn

**Abstract.** As a well-known NP-hard problem, the Three-Index Assignment Problem (AP3) has attracted lots of research efforts for developing heuristics. However, existing heuristics either obtain less competitive solutions or consume too much running time. In this paper, a new heuristic named Approximate Muscle guided Beam Search (AMBS) is developed to achieve a good trade-off between solution quality and running time. By combining the approximate muscle with beam search, the solution space size can be significantly decreased, thus the time for searching the solution can be sharply reduced. Extensive experimental results on the benchmark indicate that the new algorithm is able to obtain solutions with competitive quality and it can be employed on instances with large-scale. Work of this paper not only proposes a new efficient heuristic, but also provides a promising method to improve the efficiency of beam search.

## 1  Introduction

The Three-Index Assignment Problem (AP3) was first introduced by Pierskalla [1, 2], which is a well-known NP-hard problem with wide applications, including addressing a rolling mill, scheduling capital investments, military troop assignment, satellite coverage optimization [1, 2], scheduling teaching practice [3], and production of printed circuit boards [4]. It can be viewed as an optimization problem on a 0-1 programming model:

$$\min \sum_{i \in I} \sum_{j \in J} \sum_{k \in K} c_{ijk} x_{ijk}$$

subject to

$$\sum_{j \in J} \sum_{k \in K} x_{ijk} = 1, \quad \forall i \in I$$

$$\sum_{i \in I} \sum_{k \in K} x_{ijk} = 1, \quad \forall j \in J$$

$$\sum_{i \in I} \sum_{j \in J} x_{ijk} = 1, \quad \forall k \in K$$

$$x_{ijk} \in \{0,1\}, \quad \forall i \in I, j \in J, k \in K$$

where  $I = J = K = \{1, 2, 3, ..., n\}$.

The solution of AP3 can be presented by two permutations:

$$\min \sum_{i}^{n} c_{i,p(i),q(i)}, \quad p(i), q(i) \in \pi_N$$

where $\pi_N$ presents the set of all permutations on the integer set $N = \{1,2,...,N\}$.

Due to its intractability, there have been lots of exact and heuristic algorithms proposed to solve it, including Balas and Saltzman [5], Crama and Spieksma [6], Burkard and Rudolf [7], Pardalos and Pitsoulis [8], Voss [9], Aiex, Resende, Pardalos, and Toraldo [10], Huang and Lim [11], Jiang, Xuan, and Zhang [12]. Among these algorithms, LSGA proposed by Huang and Lim [11], and AMGO proposed by Jiang, Xuan, and Zhang [12] perform better than the other heuristics. LSGA is a hybridization of a genetic algorithm and a local search algorithm, which projects AP3 into AP2. It can obtain a solution within quite a short time. However, on difficult instances, LSGA might not perform well in terms of the solution quality. AMGO includes two phases, the sampling phase and the global optimization phase. In the sampling phase, a number of local optima are produced, and make up a smaller search space. In the second phase, a recursive procedure is employed on the new search space to find a better solution. In contrast to LSGA, AMGO can obtain better solutions with high quality. However, the running time of AMGO is intolerable on large instances. It would be ideal to achieve a good trade-off between solution quality and running time.

To tackle the challenges in balancing solution quality and running time, we propose a new heuristic named Approximate Muscle guided Beam Search (AMBS), which combines two phases. In the first phase, namely the sampling phase, a multi-restart local search algorithm is used to generate the approximate muscle. The muscle is the union of optimal solutions. Obviously, if we could get the muscle, it would be quite easy for us to obtain the optimal solution because the search space is significantly reduced. However, it has been proved that there is no polynomial time algorithm to get the muscle of an AP3 instance [12]. We can use an alternative method by using the union of local optima to approximate the muscle. The search will be restricted in the approximate muscle so that the search space is reduced dramatically. In the latter phase, namely the search phase, beam search is employed to obtain a solution on the approximate muscle. Beam search is an adaptation of breadth-first search, in which only the most promising nodes are selected for further branching at each level rather than all the nodes. Because it just searches the promising branches, the running time will not be intolerable. By combining the approximate muscle and beam search, we can obtain solutions with relatively high quality in a short time. Experimental results on the standard AP3 benchmark indicate that the new heuristic is able to obtain solutions with competitive quality and it can be employed on instances with large-scale. In terms of solution quality, the solutions obtained by AMBS are better than LSGA and not worse than the pure beam search, while in terms of running time, AMBS is able to deal with large instances that AMGO and the pure beam search cannot.

The rest of this paper is organized as follow. In Section 2, we shall give a review of the muscle and beam search. In section 3, we shall propose the framework of AMBS. Experiment results will be reported in Section 4. Finally, the conclusion of this paper will be presented in Section 5.

## 2 Muscle and Beam Search

In this section, we present the two conceptions related to this paper, namely the muscle and beam search. For each concept, we first briefly review its related work and then present its details.

### 2.1 Muscle

The proposition of the concept muscle is inspired by the backbone. The backbone is an important tool for NP-hard problem, which means the shared common parts of optimal solutions for an instance. Lots of heuristic algorithms have been proposed with the concept backbone. For example, Schneider used the intersection of local optima as the approximate backbone to solve the traveling salesman problem (TSP) [13], Zhang and Looks developed a LK algorithm guided by backbone for the traveling salesman problem [14], Xuan, Jiang, Ren, and Luo presented a backbone-based multilevel algorithm to solve the large scale next release problem [15].

In contrast to the backbone, the muscle is the union of optimal solutions. It was first proposed by Jiang, Xuan, and Zhang in 2008 [12]. Some efficient algorithms have been proposed using the muscle. For example, Jiang and Chen developed an algorithm for solving the Generalized Minimum Spanning Tree problem with the muscle [16]. Obviously, if the muscle could be obtained, the search space for an instance would be decreased sharply. However, Jiang has proved that there is no polynomial time algorithm to obtain the muscle for AP3 problem under the assumption $P \neq NP$. It is intractable to obtain a fraction of the muscle as well [12].

Now that the muscle cannot be obtained directly, there are some other ways to approximate the muscle for AP3. The "big valley" structure appears in lots of problem models, including the travelling salesman problem (TSP) [17], the graph partitioning problem [18], the job shop scheduling [19], etc. This structure suggests that the clusters formed by a lot of local optima will be around the optimal solutions. The experiments conducted by Jiang indicate that the probability that the union of local optima contains the optimal solution increases with the growth of the number of local optimum, while the size of the union increases slower [12]. Hence, we can use the union of local optima to approximate the muscle to decrease the search space, and we can name the union as the approximate muscle.

However, the approximate muscle varies with the difficulty of instances. On difficult instances, the local optima have less common triples, thus, the size of the approximate muscle will be larger. On easy instances, the local optima have more common triples, thus, the size of the approximate muscle will be smaller. Although the search space has been decreased by using the approximate muscle, the search space for AP3 problem still increases exponentially. If an exhaustive search algorithm is employed on the approximate muscle of a large instance, the running time will be unacceptable.

## 2.2 Beam Search

Beam search is a widely-used heuristic algorithm. For example, Cazenave combined Nested Monte-Carlo Search with beam search to enhance Nested Monte-Carlo Search [20], López-Ibáñez and Blum combined beam search with ant colony optimization to solve the travelling salesman problem (TSP) with time windows [21].

Beam search can be viewed as an adaptation of branch-and-bound search, or an optimization of best-first search. By branching the most promising nodes at each level rather than the whole search tree, beam search can find a solution with relatively high quality within practical time and memory limits.

The standard version of beam search builds its search tree using breadth-first search. At each level of the search tree, a heuristic algorithm is employed to estimate all the successors, and the nodes are sorted in the order of the heuristic cost, then a predetermined number of best nodes are stored, while the others are pruned off permanently. The predetermined number is called the beam width. By varying the beam width, beam search varies from greedy search to a complete breadth-first search. When the beam width equals to 1, it becomes a greedy search. When there is no limit to the beam width, it becomes a breadth-first search. By limiting the beam width, the complexity of the search becomes polynomial instead of exponential. We call the standard version of beam search as the pure beam search, in order to distinguish it with AMBS.

## 3 Approximate Muscle Guided Beam Search for AP3

In this section, we introduce the details of our algorithm AMBS. As mentioned before, AMBS includes two phases, the sampling phase and the search phase. We will first present the framework of our algorithm, and then show the details of each phase in the following subsections.

### 3.1 AMBS for AP3

The framework of the algorithm is shown in Algorithm 1. There are three inputs for AMBS, i.e. the AP3 instance, denoted by $AP3(I, J, K, c)$, the number of sampling, denoted by $k$, and the beam width, denoted by $width$. The output of this algorithm is the solution of the input instance, denoted by $s^*$. The instance of AP3 is stored in a three-dimensional array, in which an element represents cost $c_{ijk}$, as introduced in section 1. The solution is stored in two arrays, which record the two permutations.

As shown in Algorithm 1, AMBS has two phases. In the beginning of the search phase, the order of search level for beam search is sorted. This is a preprocessing for beam search to get the map of level of the search tree and the index of $I$ of the approximate muscle. A search level means an index of $I$. For example, the level $order[i]$ of the search tree has the nodes which corresponds to the triples whose index $I$ is $i$ in the approximate muscle. More details about building the search tree is introduced in section 3.3. The order of search level is sorted in ascending order by the number of triples at the corresponding index, i.e. the number of nodes in each level in the search tree. In this way, there will be fewer nodes in the higher level. When calculating the lower bound of each branch, which is employed in beam search frequently, more time will be consumed when the branch is at the higher level. Thus, after the sorting, beam search will take less searching time.

In the following subsections, we will discuss the details of two phases, respectively.

---

**Algorithm 1**: AMBS for AP3
**Input**: AP3 instance $AP3(I,J,K,c)$, $k$, $width$
**Output**: solution $s^*$
**Begin**
//the sampling phase
  (1)    obtain the approximate muscle $a\_muscle$ and a solution $s'$ as the upper bound with the algorithm $GenerateAM\ (AP3(I,J,K,c),k)$;
//the search phase
  (2)    sort the search order of the approximate muscle and get the $order$;
  (3)    obtain the solution $s^*$ with the beam search algorithm $BS(a\_muscle, width, s', order)$;
**End**

---

## 3.2 Approximate Muscle for AP3

In the first phase, namely the sampling phase, the muscle is obtained for the further searching. Although the muscle cannot be obtained directly, we can use the union of local optima to approximate the muscle. Thus, the main purpose of this phase is to obtain different local optima and get the union of them.

The detail of generating the local optima is shown in Algorithm 2. The inputs of Algorithm 2 are an instance of AP3 problem and the number of sampling times of sampling. The instance is denoted by $AP3(I,J,K,c)$, and the number of sampling time is denoted by $k$. The output of Algorithm 2 includes the approximate muscle of the input instance, which is denoted by $a\_muscle$, and the best solution obtained in the sampling phase, which is used as the upper bound of the search phase. The approximate muscle is stored in a three-dimensional array, the same as the storage of an instance, where the cost is the same value as the instance if this cost is sampled, or infinite if the corresponding cost is not sampled.

In the beginning, the approximate muscle is initialized as an empty set, which means all the values of the approximate muscle is infinite (line(1)). Then $k$ local optima are obtained, and make up the approximate muscle (line(2)-line(15)). A random feasible solution is generate by swapping the order of permutations $p$ and $q$ randomly (line(3)-line(11)). Then a local search algorithm is applied to the random solution to obtain a local optimum (line(12)). The local search algorithm we use here is the Hungarian local search, which is proposed by Huang and Lim [11]. Once a local optimum is obtained, it is added to the approximate muscle(line(13)), and compared to the best local optimum ever found, then the better solution will be kept as the best local optimum (line(14)).

---

**Algorithm 2**: GenerateAM (Generate Approximate Muscle)
**Input**: AP3 instance $AP3(I,J,K,c)$, $k$
**Output**: $a\_muscle$, solution $s'$
**Begin**
(1)    $a\_muscle = \varnothing$
(2)    for $counter = 1$ to $k$ do
(3)        for $i = 1$ to $n$ do
(4)            $p[i] = i$, $q[i] = i$;
(5)        for $i = 1$ to $n$ do
(6)            let $j$ be a random integer between 1 and $n$;
(7)            swap $p[i]$ and $p[j]$;
(8)        for $i = 1$ to $n$ do
(9)            let $j$ be a random integer between 1 and $n$;
(10)           swap $q[i]$ and $q[j]$;
(11)       let $s = \{(i, p[i], q[i]) \mid 1 \leq i \leq n\}$;
(12)       obtain a local optimum $s_{local}$ by applying the local search to $s$;
(13)       $a\_muscle = a\_mucle \cup s_{local}$;
(14)       if $c(s_{local}) < c(s')$ then $s' = s_{local}$
**End**

---

### 3.3 Beam Search for AP3

In the second phase, namely the search phase, we use beam search on the approximate muscle to find a better solution.

Before the introduction of beam search for AP3, we will first present how we build the breadth-first search tree for AP3 problem. An instance of AP3 can be represented as a three-dimensional matrix. First, the matrix is divided into $n$ layers based on the index $I$, i.e. the triples with the same index $I$ are in the same layer. Each layer corresponds to a level in the search tree. For example, an instance with the size of 4 is divided into 4 layers. Select a layer to be level 1 of the search tree (level 0 of the search tree is the root, which represents nothing). Thus, there are 16 (4*4) nodes in level 1. Then another layer is selected to build level 2 of the search tree. Since one triple has been determined in level 1, the triples in the same row and column will no longer be considered, there are 9 (3*3) successors of each node in level 1. Thus, 144 (9*16) nodes are in level 2 in all. In the same way, level 3 and level 4 are built. If we use the muscle rather than the whole instance to build the search tree, only the triples in the muscle are taken into consideration, rather than all the triples in each layers. Every node in the search tree represents a triple of the instance, except the root node. A path from the root to a leaf in the search tree represents a solution of the instance.

The detail of beam search for AP3 is presented in Algorithm 3. The inputs of Algorithm 3 are the approximate muscle $a\_muscle$, the beam width $width$, the best local $s'$, and the search order $order$. The output of Algorithm 3 is the solution $s^*$ of this AP3 instance.

**Algorithm 3**: BS (Beam Search)
**Input**: $a\_muscle$, $width$, solution $s'$, $order$
**Output**: solution $s^*$
**Begin**
(1)   for every $level$ based on $order$ in the search tree do
(2)      for every $candidate$ do
(3)         for every triple $(i,j,k) \in candidate$ do
(4)            $fp[j] = true$, $fq[k] = true$;
(5)         for every triple $(order[level], j, k) \in a\_muscle$ do
(6)            if $fp[j] = false$ and $fq[k] = false$ then
(7)               generate the sub-problem;
(8)               calculate the lower bound of the branch;
(9)      sort the branches of all the $candidate$;
(10)     for $i = 1$ to $width$ do
(11)        if lower bound of the branch $< c(s')$ then
(12)           this branch belongs to the new $candidates$;
(13)        else break;
(14)  employ the local search algorithm on every $candidate$ and choose the best to be the solution $s^*$
**End**

As mentioned before, beam search builds its search tree using breadth-first search. The nodes in each level of the search tree represent the triples with the same index $I$. The algorithm searches the tree level by level first (line(1)-line(13)). A candidate represents an incomplete solution, or the nodes on an incomplete search path, i.e. part of the triples in a solution, or a branch stored to be searched. When the search comes to a certain level, the lower bounds of all the successors of each candidate are generated (line(2)-line(8)). For each candidate, the determined triples are recorded first using the array $fp$ and $fq$. In this way, the constraints of AP3 can be guaranteed, and the sub-problem can be obtained (line(3)-line(4)). Then the lower bounds of successors of this candidate are calculated (line(5)-line(8)). The lower bound of a successor includes three parts: the value of candidate, which means the sum of the triples' cost in the candidate, the cost of the triple relevant to the successor, and lower bound of the sub-problem. The sub-problem is the approximate muscle without the layers containing the determined triples. The algorithm that we use here for calculating the lower bound is the Projection method followed by a Hungarian algorithm, which is proposed by Kim et al. [22]. All the successors of each candidate are sorted in ascending order according to their lower bounds (line(9)). If some successors have the same lower bound, the successor whose predecessor (i.e. the candidate) has the smaller average lower bounds of all the successors will rank more forward. In the end, at most *width* successors are kept to be the new candidates, and all the new candidates must have a smaller lower bound than the cost of solution $s'$ (line(10)-line(13)). After the search in the search tree, a local search algorithm is employed to the remaining candidates, which have become the solutions of the instance. Then the best candidate is chosen to be the solution $s^*$. If all the candidates have higher lower bound or cost than the cost of solution $s'$, there will be no candidate remained, the solution $s'$ will be chosen as the $s^*$ (line(14)).

Since the approximate muscle is stored in the same way as an instance, beam search algorithm can be used to solve AP3 problem independently.

## 4   Experimental Result

In this section, we first show the result of parameter tuning. Then we present the experimental results of our algorithm on the benchmark compared with LSGA, AMGO, and the pure beam search. The codes are implemented with C++ under windows 7 using visual studio 2010 on a computer with Intel Core i3-M330 2.13G.

### 4.1   Parameter Tuning

Two parameters are used in AMBS, the number of sampling and the beam width. We determine the number of sampling as 1000, the same value in AMGO [12]. As for the beam width, we test different beam widths {100, 200, 300, 400} on some instances from Balas and Saltzman Dataset (see Section 4.2) and Crama and Spieksma Dataset (see Section 4.3). The instances of Balas and Saltzman Dataset we used are the instances of large size, i.e. bs_14_x.dat, bs_18_x.dat, bs_22_x.dat, and bs_26_x.dat. Here x means 1 to 5, since there are 5 instances of each size. The instances of Crama and Spieksma Dataset we used are the first instance of each type, each size, i.e. 3DA99N1 (Type I, size 33), 3DA198N1 (Type I, size 66), 3DIJ99N1 (Type II, size 33), 3DI198N1 (Type II, size 66), 3D1299N1 (Type III, size 33), and 3D1198N1 (Type III, size 66).

Table 1 shows the result of our parameter tuning experiment. We run the algorithm 10 times on the instances of the Balas and Saltzman Dataset with each beam width, while run it once on the instances of Crama and Spieksma Dataset with each beam width since the instances of Crama and Spieksma Dataset are much larger but the result of each instance varies little. The value of Balas and Saltzman Dataset in the table is the average value of each size. The experimental result indicates that, on most of the instances, the quality of the solution rises with the increase of the beam width, while the running time grows, too. The running time grows linearly with the increase of the beam width. Note that the running time of 3DA99N1, 3DA198N1 and 3D1299N1 keeps the same in different beam widths. The approximate muscle space for 3DA99N1 is so small that when the beam width is 100, beam search has become a complete search. As for 3DA198N1, after the sort for search level, the number of node on the higher level is 1, while the search on the lower level is quite quick, thus the running time varies little when the beam width changes. When testing 3D1299N1, the upper bound found in the sampling phase is the same as the smallest lower bound of the first level in the search tree, thus the upper bound is the optimal solution of the instance and all the branches are pruned, the search terminates. Among these instances, the running time of 3DI198N1 is the longest. When the beam width is 300, the running time is about 20 minutes. In order to balance the quality of the solution and the running time, we determine the beam width as 300 in the rest of experiments.

**Table 1.** Beam Width Tuning

| Instance Id | Width=100 | | Width=200 | | Width=300 | | Width=400 | |
|---|---|---|---|---|---|---|---|---|
| | Cost | Time (sec) | Cost | Time (sec) | Cost | Time (sec) | Cost | Time (sec) |
| BS_14_x | 10 | 0.74 | 10 | 1.07 | 10 | 1.31 | 10 | 1.72 |
| BS_18_x | 6.86 | 2.48 | 6.66 | 4.25 | 6.48 | 5.97 | 6.46 | 7.82 |
| BS_22_x | 4.86 | 6.75 | 4.62 | 12.00 | 4.34 | 17.13 | 4.34 | 22.45 |
| BS_26_x | 2.74 | 15.01 | 2.34 | 27.05 | 2.1 | 39.16 | 2.08 | 51.01 |
| 3DA99N1 | 1608 | 7.01 | 1608 | 7.39 | 1608 | 7.24 | 1608 | 7.16 |
| 3DA198N1 | 2662 | 62.05 | 2662 | 65.00 | 2662 | 63.15 | 2662 | 64.34 |
| 3DIJ99N1 | 4797 | 16.59 | 4797 | 18.53 | 4797 | 20.33 | 4797 | 22.11 |
| 3DI198N1 | 9685 | 479.66 | 9684 | 863.94 | 9684 | 1219.82 | 9684 | 1566.40 |
| 3D1299N1 | 133 | 1.70 | 133 | 1.75 | 133 | 1.73 | 133 | 1.73 |
| 3D1198N1 | 286 | 169.37 | 286 | 276.14 | 286 | 383.95 | 286 | 573.86 |

### 4.2 Balas and Saltzman Dataset

This dataset is generated by Balas and Saltzman [5] which contains 60 instances with size of 4, 6, 8, ..., 24, 26. For each size, five instances are generated randomly with the cost between 0 and 100.

Table 2 shows the experimental result on this dataset. Each row represents the average cost of the same size. The column "Opt." is the optimal solution reported by Balas and Saltzman [5]. The column "LSGA" is the result reported in Huang's paper [11], where a Pentium III 800MHz PC is used for the experiment. The column "AMGO" is the results of the program implemented according to Jiang's paper [12]. Interestingly, the average cost of size 26 is 1 in this column, which is better than the optimal solution reported by Balas and Saltzman. Because the average of any five integers cannot be 1.3, we think than it may be a typo in Balas's paper. The solutions of these five instances found in this paper are 0, 0, 2, 1, 2, respectively. The column "Beam Search" is the result of the pure beam search for AP3. The sampling phase remains to get the upper bound of an instance. The column "AMBS" is the result of our algorithm. The results of AMGO, the pure beam search and AMBS are the average cost after running the algorithm 10 times on each instance.

The result indicates that AMBS can get solutions with higher quality than LSGA. AMGO can generate the best solutions, and the running time is quite short on this dataset, because it employs a global search on the approximate muscle and the search space of the approximate is quite small. AMBS uses an incomplete search and needs to estimate the lower bound of each branch, thus, the quality of solutions is a little worse and the running time is longer than AMGO. Compared with the pure beam search, the running time of AMBS is about one-tenth of the pure beam search, but the quality of the solutions of AMBS is comparable to that of beam search. The reason is that with the introduction of the muscle, when calculating the lower bounds of all the successors of each level, there are much fewer successors in the approximate muscle.

**Table 2.** Balas and Saltzman Dataset (12*5 instances)

| n | Opt. | LSGA (PIII800) | | AMGO (i3-M330 2.13G) | | Beam Search (i3-M330 2.13G) | | AMBS (i3-M330 2.13G) | |
|---|------|------|------------|-------|------------|-------|------------|-------|------------|
|   |      | Cost | Time (sec) | Cost  | Time (sec) | Cost  | Time (sec) | Cost  | Time (sec) |
| 4  | 42.2 | 42.2 | 0     | 42.2  | 0.01  | 42.2 | 0.01   | 42.2 | 0.01  |
| 6  | 40.2 | 40.2 | 0.01  | 40.2  | 0.03  | 40.2 | 0.03   | 40.2 | 0.03  |
| 8  | 23.8 | 23.8 | 0.03  | 23.8  | 0.06  | 23.8 | 0.06   | 23.8 | 0.06  |
| 10 | 19   | 19   | 0.37  | 19    | 0.11  | 19   | 0.14   | 19   | 0.11  |
| 12 | 15.6 | 15.6 | 0.87  | 15.6  | 0.18  | 15.6 | 0.62   | 15.6 | 0.26  |
| 14 | 10   | 10   | 1.73  | 10    | 0.26  | 10   | 7.62   | 10   | 1.31  |
| 16 | 10   | 10   | 1.89  | 10.16 | 0.52  | 10   | 22.74  | 10   | 3.32  |
| 18 | 6.4  | 7.2  | 2.95  | 6.4   | 0.97  | 6.4  | 49.65  | 6.48 | 5.97  |
| 20 | 4.8  | 5.2  | 4.01  | 4.8   | 1.67  | 4.8  | 98.18  | 4.88 | 10.43 |
| 22 | 4    | 5.6  | 4.54  | 4     | 6.26  | 4.24 | 185.70 | 4.34 | 17.13 |
| 24 | 1.8  | 3.2  | 5.66  | 1.96  | 12.16 | 2.38 | 313.60 | 2.28 | 26.95 |
| 26 | 1.3  | 3.6  | 10.78 | 1     | 6.62  | 2.38 | 526.59 | 2.1  | 39.16 |

### 4.3 Crama and Spieksma Dataset

This dataset is generated by Crama and Spieksma [6]. This dataset contains three types of instance. In each types, there are three instances with the size of 33, and three instances with the size of 66.

Table 2 shows the experimental result on this dataset. The column "LSGA" is the result reported in Huang's paper [11]. The column "AMGO" is the result of AMGO implemented according Jiang's paper, too. The column "Beam Search" is the result of the pure beam search without the approximate muscle. The column "AMBS" is the result of our algorithm. AMGO, the pure beam search and AMBS are executed once, and the results of the algorithms are reported in Table 2. In the table, there are some cells with no value in it, this is because the running time is longer than 30 minutes, and we regard this time as unacceptable.

From the result, we can see that AMBS is able to run on every instance, and obtain a solution with high quality, however, AMGO and the pure beam search are not able to deal with a number of instances in this dataset because it takes too much time. The running time of LSGA is quite short, but the solution quality is high, this is because LSGA is an iterative algorithm rather than a tree search, and the instances in this dataset are easy to solve. If the instance is hard to solve, like the large instance in Balas and Saltzman Dataset, the quality of solutions of LSGA might not be that high.

**Table 3.** Crama and Spieksma Dataset. (18 instances)

| n | Instance Id | LSGA (PIII800) | | AMGO (i3-2120 3.3G) | | Beam Search (i3-2120 3.3G) | | AMBS (i3-2120 3.3G) | |
|---|---|---|---|---|---|---|---|---|---|
| | | Cost | Time (sec) | Cost | Time (sec) | Cost | Time (sec) | Cost | Time (sec) |
| 33 | 3DA99N1 | 1608 | 0.03 | 1608 | 7.60 | 1608 | 649.74 | 1608 | 7.24 |
| 33 | 3DA99N2 | 1401 | 0.11 | 1401 | 7.11 | 1401 | 1733.90 | 1401 | 6.52 |
| 33 | 3DA99N3 | 1604 | 0.11 | 1586 | 7.61 | 1604 | 1606.99 | 1604 | 7.30 |
| 66 | 3DA198N1 | 2662 | 0.55 | 2662 | 71.22 | - | - | 2662 | 63.15 |
| 66 | 3DA198N2 | 2449 | 0.27 | - | - | - | - | 2449 | 74.20 |
| 66 | 3DA198N3 | 2758 | 0.58 | - | - | - | - | 2758 | 82.07 |
| 33 | 3DIJ99N1 | 4797 | 0.11 | - | - | - | - | 4797 | 20.33 |
| 33 | 3DIJ99N2 | 5067 | 0.26 | - | - | - | - | 5067 | 35.95 |
| 33 | 3DIJ99N3 | 4287 | 0.26 | - | - | - | - | 4287 | 26.07 |
| 66 | 3DI198N1 | 9684 | 4.86 | - | - | - | - | 9684 | 1219.82 |
| 66 | 3DI198N2 | 8944 | 3.35 | - | - | - | - | 8944 | 929.51 |
| 66 | 3DI198N3 | 9745 | 3.09 | - | - | - | - | 9745 | 767.66 |
| 33 | 3D1299N1 | 133 | 0.01 | - | - | 133 | 3.50 | 133 | 1.73 |
| 33 | 3D1299N2 | 131 | 0.03 | - | - | 131 | 1128.17 | 131 | 3.94 |
| 33 | 3D1299N3 | 131 | 0.02 | 131 | 1.98 | 131 | 580.97 | 131 | 3.31 |
| 66 | 3D1198N1 | 286 | 0.15 | - | - | - | - | 286 | 383.95 |
| 66 | 3D1198N2 | 286 | 0.16 | - | - | - | - | 286 | 341.05 |
| 66 | 3D1198N3 | 282 | 0.23 | - | - | - | - | 282 | 329.67 |

## 5 CONCLUSION

In this paper, we propose a new heuristic named Approximate Muscle guided Beam Search (AMBS) for AP3 problem. This algorithm combines the approximate muscle and beam search. AMBS includes two phases, the sampling phase, in which the approximate muscle is obtained, and the search phase, in which beam search is employed. In this way, the solution space size of AP3 problem is decreased significantly, and the search time is reduced sharply, too. Thus, AMBS can achieve a good trade-off between the solution quality and the running time. Experimental results indicate that the new algorithm is able to obtain solutions with competitive quality, and it can be employed on large-scale instances.